%% file: main.tex
\documentclass[10pt,twocolumn,letterpaper]{article}

\usepackage[final]{cvpr}              
\usepackage{bm}
\usepackage{multirow}


\definecolor{cvprblue}{rgb}{0.21,0.49,0.74}
\usepackage[pagebackref,breaklinks,colorlinks,allcolors=cvprblue]{hyperref}

\def\paperID{5307} 
\def\confName{CVPR}
\def\confYear{2025}

\title{Modeling Thousands of Human Annotators for Generalizable Text-to-Image Person Re-identification}

\author{
    {Jiayu Jiang$^{1}$~~~~~
    Changxing Ding$^{1,2}$\thanks{Corresponding author}~~~~~
    Wentao Tan$^{1}$~~~~~
    Junhong Wang$^{1}$~~~~~
    Jin Tao$^1$~~~~~
    Xiangmin Xu$^{1}$} 
    \\
    $^1$South China University of Technology \quad 
    $^2$Pazhou Lab, Guangzhou \quad
    \\
    {\tt\small 202320111494@mail.scut.edu.cn, chxding@scut.edu.cn}
    \\
    {\tt\small \{ftwentaotan,eewjh\}@mail.scut.edu.cn, \{arjtao,xmxu\}@scut.edu.cn}
}

\begin{document}
\maketitle
\input{sec/0_abstract}    
\input{sec/1_intro}

\input{sec/2_related}

\input{sec/3_method}
\input{sec/4_experiment}
\input{sec/5_conclusion}
{
    \small
    \bibliographystyle{ieeenat_fullname}
    \bibliography{main}
}


\end{document}

%% file: sec/0_abstract.tex
\begin{abstract}
Text-to-image person re-identification (ReID) aims to retrieve the images of an interested person based on textual descriptions. One main challenge for this task is the high cost in manually annotating large-scale databases, which affects the generalization ability of ReID models. Recent works handle this problem by leveraging Multi-modal Large Language Models (MLLMs) to describe pedestrian images automatically. However, the captions produced by MLLMs lack diversity in description styles. To address this issue, we propose a Human Annotator Modeling (HAM) approach to enable MLLMs to mimic the description styles of thousands of human annotators. Specifically, we first extract style features from human textual descriptions and perform clustering on them. This allows us to group textual descriptions with similar styles into the same cluster. Then, we employ a prompt to represent each of these clusters and apply prompt learning to mimic the description styles of different human annotators. Furthermore, we define a style feature space and perform uniform sampling in this space to obtain more diverse clustering prototypes, which further enriches the diversity of the MLLM-generated captions. Finally, we adopt HAM to automatically annotate a massive-scale database for text-to-image ReID. Extensive experiments on this database demonstrate that it significantly improves the generalization ability of ReID models. Code is available at \url{https://github.com/sssaury/HAM}.
\end{abstract}

%% file: sec/1_intro.tex
\section{Introduction}
\label{sec:intro}
Text-to-image person re-identification (ReID) \cite{wang2020vitaa,ding2021semantically,shao2022learning,farooq2022axm,yan2023learning, jiang2023cross,shao2023unified,yan2023learning} aims to retrieve a person of interest from a large image gallery based on textual descriptions. It plays an important role in intelligent video surveillance \cite{bukhari2023language}, public security \cite{bukhari2023language}, and social media analysis \cite{galiyawala2021person}. However, text-to-image ReID remains challenging especially in terms of the generalizability of existing models. This is primarily due to the high cost in textual annotation, which significantly affects the size of existing manually-labeled datasets.

To address this issue, recent works proposed various methods to automatically generate captions for massive-scale pedestrian images \cite{shao2023unified,yang2023towards,tan2024harnessing,zuo2023plip}. These works can be categorized into two groups. One category of methods recognize each pedestrian image's attributes and fills them into manually designed description templates \cite{shao2023unified,zuo2023plip}. The other category of methods employ off-the-shelf Multi-modal Large Language Models (MLLMs) to produce textual descriptions for each pedestrian image \cite{yang2023towards}. They also rely on manually designed description templates to enhance the diversity of the obtained textual descriptions \cite{tan2024harnessing}.

\begin{figure}[t]
  \centerline{\includegraphics[width=1.0\linewidth, height=6.5cm]{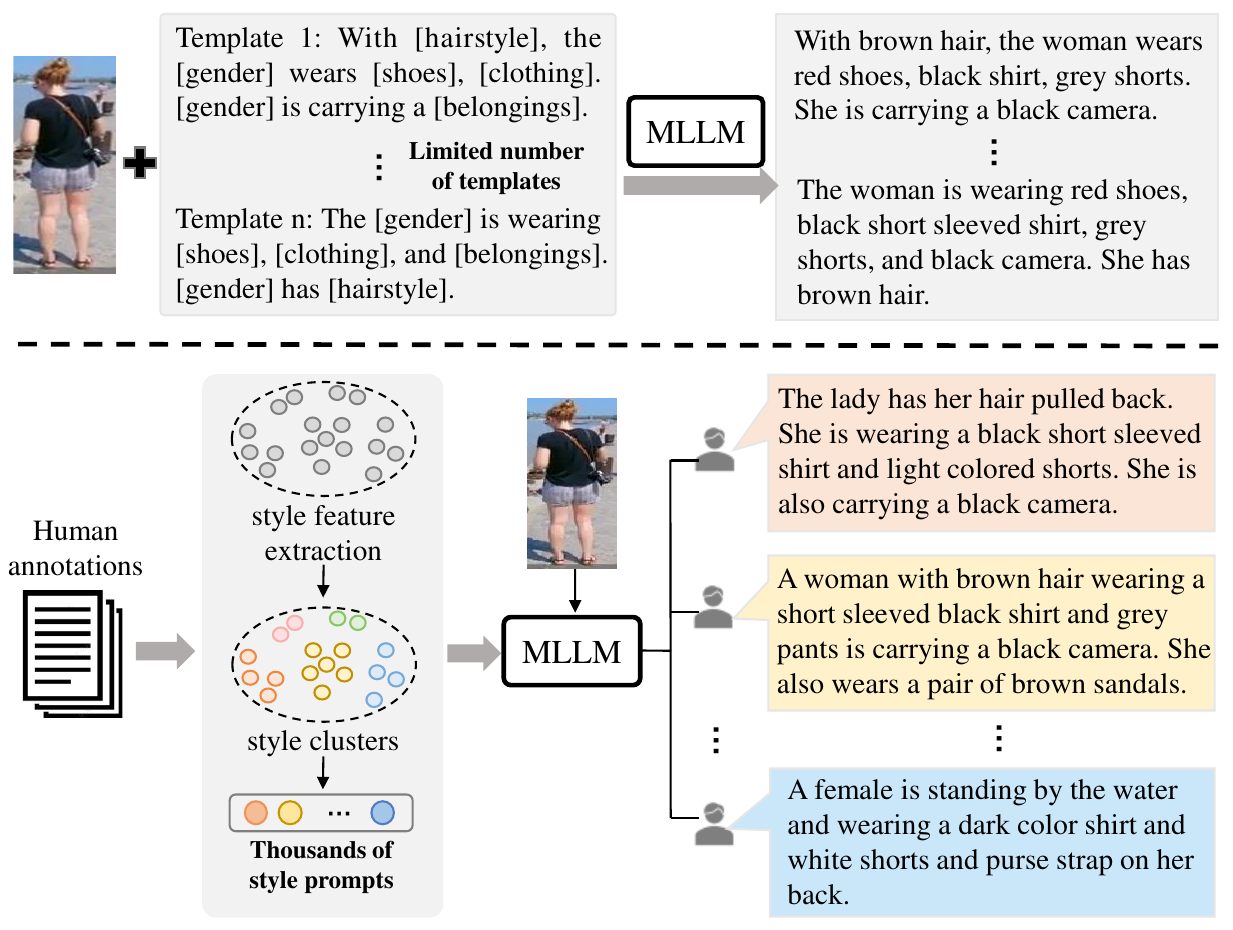}}
  \vspace{-0em}
  \caption{Illustration of different schemes for enhancing the style diversity for MLLM-generated textual descriptions. (Top) Existing works typically rely on manually designed description templates to instruct MLLM to generate style-diverse captions. (Bottom) Our approach models the description styles of thousands of human annotators, and enables the MLLM to capture these styles in a learning-based approach.}
  \label{fig:fig1}
  \vspace{-1em}
\end{figure}

However, as illustrated in Figure \ref{fig:fig1}, these templates are designed in a heuristic manner with typically limited numbers (\textit{e.g.}, 46 in \cite{tan2024harnessing} and 456 in \cite{shao2023unified}). This means that these templates can only cover limited description styles encountered in real-world applications. Moreover, these templates usually focus on sentence patterns, without considering the styles in wording for pedestrian attributes (\textit{e.g.}, the placeholder like [hairstyle] fails to capture the difference between [straight shoulder-length black hair] and [black hair] in style). In comparison, the description styles from human annotators are much richer as different annotators have distinct preferences. The main limitation of human annotation lies in the high cost in scaling on large-scale image data. Therefore, it is highly desirable to have an approach that learns the description styles of human annotators and prompts MLLMs to generate captions for massive-scale images according to these learned styles.

Herein, we propose a Human Annotator Modeling (HAM) approach to enable MLLMs to mimic the description styles of thousands of human annotators. We achieve this based on the human-annotated textual descriptions in existing text-to-image ReID databases. First, we extract a style feature from each pedestrian description. Specifically, we identify the words in a textual description that describe pedestrian attributes, \textit{e.g.}, clothing category, color, and age. Then we replace these words with vague ones that are generic across identities using an LLM. We feed the obtained textual description into the text encoder of the CLIP model \cite{radford2021learning} and the output is utilized as the style feature. Second, we perform clustering on the style features of all human-annotated textual descriptions to group textual descriptions of similar styles. Third, we employ a prompt to represent the description style of one cluster and apply prompt learning to optimize all prompts, which enables the MLLM to follow different style prompts. In contrast to using pre-defined templates that ignore wording styles, prompt learning from complete descriptions can help us obtains comprehensive style information. Finally, by enlarging the number of clusters, we can model thousands of human annotators using a single MLLM.

In our HAM method, an approach is required to group similar style features into the same cluster. However, popular clustering methods, \textit{e.g.}, KMeans \cite{macqueen1967some} and DBSCAN \cite{ester1996density} are based on the prototype-sample or sample-sample distances, leading to an uneven distribution of the obtained cluster centers. This contradicts with our motivation that the cluster centers should capture as rich human annotation styles as possible. To handle this problem, we first define a style feature space for textual descriptions. In this space, we uniformly sample cluster centers to capture the most diverse description styles. Then, we assign a fixed number of samples that are the closest to each cluster center to obtain all clusters. We refer to this methods as Uniform Prototype Sampling (UPS). With this method, we capture a wider range of human annotations styles and finally enrich the diversity of the MLLM-generated captions.

To the best of our knowledge, the HAM is the first approach to learn and mimic different description styles of human annotators for the text-to-image person ReID task. By modeling thousands of human annotators, we enable the MLLM to generate captions of diverse styles and thereby considerably improve the quality of our automatically annotated text-to-image ReID database. Extensive experimental results demonstrate that our database significantly improves the generalizability of text-to-image ReID models.


%% file: sec/2_related.tex
\section{Related Work}
\textbf{Text-to-Image ReID.} Existing text-to-image ReID methods mainly solve two key problems: the way to align multi-modal features and the way to obtain generalizable representations via pre-training.

Early approaches to the first key problem usually project the holistic visual and textual features in a shared feature space to reduce their modality gap \cite{chen2018improving,zhang2018deep,li2017person,yan2023learning,wang2019language,shu2022see,wang2016learning,zhang2018deep,aggarwal2020text,wu2023refined}. However, due to the high intra-class and low inter-class variations in pedestrian appearance and textual descriptions, it is difficult for the holistic features to capture the subtle difference between pedestrian identities, which leads to unreliable cross-modality matching. To address this issue, some works \cite{lee2018stacked,niu2020improving,jing2020pose,wang2022caibc,chen2022tipcb,gao2021contextual,wang2022look,tang2022learning,li2023knowledge} introduced part-level feature alignment methods, which are divided into two types: explicit and implicit alignment strategies. The explicit alignment strategies \cite{wang2020vitaa,ding2021semantically} first leverage external tools to detect pedestrian parts from images and textual phrases. Then, they employ additional networks to achieve part-level cross-modality feature alignment. In comparison, the implicit alignment strategies \cite{farooq2022axm,shao2022learning,yang2023towards,jiang2023cross} abandon external tools and adopt various regularization strategies to automatically establish the correspondence between noun phrases and specific pedestrian body-parts. Despite the success of these methods in improving the in-domain performance of ReID models, their generalization ability to unseen data domains is limited due to the small size of manually labeled datasets.

Recently, many methods have been proposed to handle the second key problem. They typically design various strategies to automatically caption pedestrian images and then obtain massive-scale image-text pairs. Then, they leverage these large databases to train generalizable text-to-image ReID models. These methods can be further categorized into two groups. The first group recognizes each pedestrian image's attributes and then inserts them into pre-defined templates to obtain complete descriptions. For instance, Shao et al. \cite{shao2023unified} leveraged the CLIP model \cite{radford2021learning} for attribute recognition and designed 456 templates in a heuristic manner. Zuo et al. \cite{zuo2023plip} utilized the CUHK-PEDES \cite{li2017person}  and ICFG-PEDES \cite{ding2021semantically}  datasets to train an image captioner that simultaneously predicts pedestrian attributes and generates complete textual descriptions. These attribute predictions are also embedded into templates to produce image captions. The second group of methods directly employ advanced MLLMs \cite{yin2023survey,bai2023qwen,chen2023shikra,li2022blip} to generate captions. For example, Yang et al. \cite{yang2023towards} employed a diffusion model \cite{rombach2022high} to synthesize pedestrian images and then utilized BLIP-2 \cite{li2022blip} to generate textual descriptions for these images. However, it does not take the textual diversity into consideration. To handle this problem, Tan et al. \cite{tan2024harnessing} designed 46 description templates in a heuristic manner with the help of ChatGPT \cite{ChatGPT}, by which they enrich the instructions to MLLM and prompt it to generate captions with diverse sentence patterns. However, the limited number of templates and their heuristic design still affect the diversity of the obtained captions.

\textbf{Diversity Enhancement of the MLLM Outputs.} Due to the high cost in manual annotations, utilizing MLLMs as a tool to automatically produce text annotations has become popular in large-scale dataset construction. However, MLLMs usually exhibit noticeable description preferences for certain content or styles, motivating researchers to address the homogeneity in MLLM-generated data. Existing approaches can be divided into two categories: prompt engineering and external knowledge-based methods. 

For the first category, Yu et al. \cite{yu2024large} enhanced the textual diversity by listing image attributes to describe in their prompts, which allows for varied outputs by altering the attribute instances. Chan et al. \cite{chan2024scaling} adopted a character-driven approach to simulate questions from different persons, resulting in a diverse range of synthesized perspectives. For the second category, Zhu et al. \cite{zhu2023chatgpt} leveraged ChatGPT \cite{ChatGPT} to generate questions from different perspectives, instructing the MLLMs to focus on different elements in the images and therefore producing a large amount of textual descriptions. Besides, Divekar et al. \cite{divekar2024synthesizrr} introduced diversity through retrieval augmentation. They incorporated the retrieved information from external databases into the text generation  process, resulting in diverse outputs. However, these methods primarily focus on generating content-diverse outputs for MLLMs, with little attention paid to enhancing the diversity of description styles for the same semantic content.

In this paper, we focus on enhancing the diversity in captioning styles of MLLMs for the same semantic content. This models the intra-class variations of textual descriptions, enabling us to obtain a large-scale and high-quality dataset that is further used for training generalizable text-to-image ReID models.

%% file: sec/3_method.tex
\section{Method}

\begin{figure*}[t]
\centering
\includegraphics[width=1.0\linewidth]{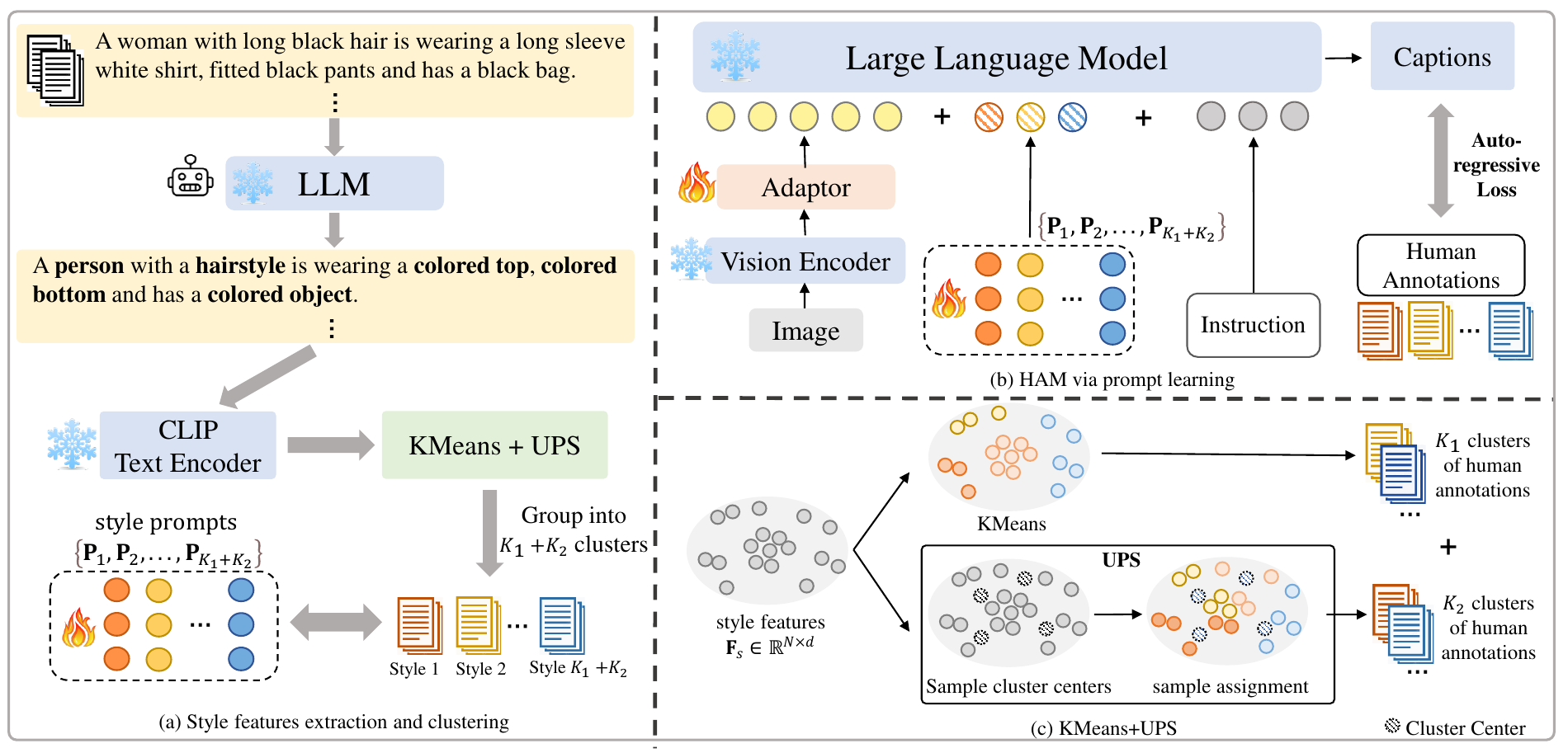}
\vspace{-2em}
\caption{Overview of our framework. In HAM, we first extract style features from human annotations and then perform clustering on these features, as illustrated in (a). This allows us to group human annotations into $K_1+K_2$ clusters, each of which contain descriptions with similar styles. We employ $K_1+K_2$ prompts to represent these description styles. (b) illustrates that we employ prompt learning to model the description preference of different annotators with MLLM. To fully explore the diverse styles in human annotations, we propose UPS, a clustering method that complements KMeans as shown in (c). }
\label{fig:framewrok}
\vspace{-1em}
\end{figure*}

The overview of our framework is illustrated in Figure \ref{fig:framewrok}. Section \ref{sec:HAM} introduces our Human Annotator Modeling (HAM) method, which leverages learnable prompts to model human annotator preferences and simulate the generation of style-diverse captions by an MLLM. Section \ref{sec:UPS} introduces the Uniform Prototype Sampling (UPS) method to complement the clustering strategy used in HAM, by which we extract a broader range of human styles and further enhance the diversity of MLLM-generated captions. Finally, in Section \ref{sec:reid model}, we use the automatically annotated dataset obtained from the above steps to train generalizable text-to-image ReID models.

\subsection{Human Annotator Modeling} \label{sec:HAM}
Due to the inherent bias, an MLLM tends to generate captions with similar sentence patterns and wording styles \cite{tan2024harnessing}. In comparison, human annotators present rich description style variations as they have distinct preferences. This inspires us to leverage an MLLM to imitate a large amount of human annotation preferences.


\textbf{Style Feature Extraction and Clustering.} To model the preference of one human annotator, the most intuitive way is to collect captions labeled by this annotator. However, this may be impractical due to the following two reasons. First, the annotator information is usually missing in existing datasets, which means we cannot collect captions labeled by each annotator. Second, an annotator may present different description preferences for different images. Therefore, we propose to extract a style feature vector from each human textual description and adopt clustering to identify descriptions with similar styles. 

To extract style features, we need to remove information that reflects the identity of one pedestrian from the textual description. To achieve this, we first identify the words in a caption that describe pedestrian attributes, \textit{e.g.}, clothing category, color, age, hairstyle, and gender. We then design instructions to prompt an LLM (Qwen2.5-7B \cite{team2024qwen2} in our experiments) to replace these words with vague ones that are generic across identities. Moreover, to better align the LLM’s output with our expectations, we provide some in-context examples in the instruction, which is detailed in the supplementary material. Finally, the obtained new textual description keeps the original description style but without words reflecting the pedestrian’s identity.

Next, we feed the obtained textual description into the text encoder of the CLIP model \cite{radford2021learning} and adopt the output holistic feature as the description’s style feature. Intuitively, textual descriptions with similar style features are more likely to belong to the same description preference, while those with dissimilar style features represent distinct preferences. Therefore, we perform clustering, \textit{e.g.}, the KMeans method \cite{macqueen1967some}, on the style features of all human-annotated textual descriptions to group them into $K_1$ clusters. After that, we attach a pseudo style label to each description based on their assigned cluster. Finally, we construct a dataset of pedestrian image-text pairs with style labels: $\mathcal{D}=\{{(\boldsymbol{x}_i, \boldsymbol{y}_i, {s}_i)}\}^N_{i=1}$
, where $\boldsymbol{x}_i$, $\boldsymbol{y}_i$ and $s_i$ represent the pedestrian image, the human-annotated textual description, and the style label for the $i$-th sample, respectively. We employ $s_i\in \left \{ 1,2,...,K_1 \right \} $ to label the description preference of one virtual human annotator.

\textbf{HAM via Prompt Learning.} In the following, we employ prompt learning popular in Vision-Language Models (VLMs) \cite{wang2023controllable, lei2024ez, wu2024cascade, cao2024promoting} to model the description preference of different annotators. Specifically, for textual descriptions in the same cluster, we employ a learnable prompt to represent their styles, resulting in a total of $K_1$ prompts. These style prompts are denoted as $\left \{ \mathbf{P}_1, \mathbf{P}_2,..., \mathbf{P}_{K_1}\right \} $ and initialized randomly. We optimize each of them with samples in the corresponding cluster. Each of these prompts is concatenated with image and text tokens at the input of the LLM component of MLLM as follows:
\begin{equation}\label{Ins}
    Ins = \left [ \mathbf{V},\mathbf{P}_i,\mathbf{T}  \right ],\mathbf{P}_i\in \mathbb{R}^{M\times D},  
\end{equation}
where $M$ is the number of tokens in each learnable prompt. $\mathbf{V}\in \mathbb{R}^{M_V\times D}$ and $\mathbf{T}\in \mathbb{R}^{M_T\times D}$ represent the image and text tokens, respectively. The instruction to prompt the MLLM to generate textual descriptions for a pedestrian image is as follows:

\emph{``Describe the person in the image without referring to the background or any additional context. Focus on their physical appearance, including their hairstyle, clothing, shoes and any noticeable accessories."}

To preserve the knowledge contained in the MLLM and avoid it overfitting to the human-annotated training data, we freeze all parameters in the LLM and the vision encoder. We only optimize the $K_1$ learnable prompts along with the vision-language adaptor \cite{team2024qwen2, liu2024llava}. We fine-tune the latter to better accommodate the concatenated style prompts, which is proved to be helpful in our experiments. Therefore, our fine-tuning strategy requires only small computing resources and memory usage, while maximizing the retention of the MLLM’s original capabilities and generalization power. We adopt a standard auto-regressive loss \cite{vaswani2017attention} for fine-tuning:
\vspace{-0.8em}
\begin{equation}\label{HAM loss}
	{\cal L}_{\text{HAM}} = - {\mathbb E}_{ \left(\boldsymbol{x}_i, \boldsymbol{y}_i, {s}_i \right) \in {\cal D}} \left[ \sum_{m} \text{log}p \left( {\boldsymbol y}_{{{i}},m}| {\boldsymbol x}_i, {\bf P}_{ s_{i}}, {\boldsymbol y}_{{i},<m} \right) \right],
\end{equation}
where ${\boldsymbol y}_{{i},m}$ denotes the $m$-th token in  $\boldsymbol{y}_i$, and $p \left( {\boldsymbol y}_{{{i}},m}|\cdot \right) $ represents the probability of the model to predict ${\boldsymbol y}_{{i},m}$.

It is worth noting that in each iteration, one sample in the ${s}_i$-th cluster is only used to update parameters in $\mathbf{P}_i$ and the adaptor. This allows us to learn a unique style prompt for each cluster, and each prompt guides the MLLM to generate captions of one unique style.

\subsection{Uniform Prototype Sampling} \label{sec:UPS}
In HAM, we adopt a clustering method to group textual descriptions with similar style  features into the same cluster. However, popular clustering methods, \textit{e.g.}, KMeans \cite{macqueen1967some} and DBSCAN \cite{ester1996density}, tend to result in an uneven distribution of cluster centers. This is because they rely on prototype-sample distance or sample-sample distance for clustering, which means clusters mainly distribute in areas with dense samples. This phenomenon restricts the ability of HAM to model wider range of human annotation preferences. To address this, we propose the Uniform Prototype Sampling (UPS) method, by which we obtain clusters that distribute evenly in our defined style feature space.

\textbf{Style Feature Space.} Research works in paraphrasing \cite{hosking2021factorising} usually disentangle the textual feature space into semantic and style spaces. Inspired by these works, we propose the definition of style feature space to model the underlying distribution of style features of all human descriptions in the dataset. We assume the description preferences vary gradually between human annotators and therefore their style features distribute in a continuous space. Following \cite{tan2023style}, we define the range of this style feature space by calculating the mean and standard deviation of each element in the style features $\textbf{F}_s \in \mathbb{R}^{N \times d}$ from the dataset $\mathcal{D} $:
\vspace{-0.5em}
\begin{equation}
    \bm \mu _{{s}}=\frac{1}{N} \sum_{n=1}^{N}\mathbf{F}^{n}_s,
\end{equation}
\vspace{-0.5em}
\begin{equation}
    \bm\sigma _{ s}=\sqrt{\frac{1}{N}\sum_{n=1}^{N}\left ( \mathbf{F}^{n}_s-\bm\mu _{s}   \right ) ^2  }, 
\end{equation}
where $d$ is the dimension of each style feature, and $\mathbf{F}^n_s$ is the $n$-th vector in $\mathbf{F}_s$.

Finally, the range of this style feature space is represented as $\left [ \bm\mu _{ s}-\beta * \bm\sigma _{ s}, \bm\mu _{ s}+\beta * \bm\sigma _{ s} \right ] $, where $\beta$ is a hyper-parameter. We then uniformly sample $K_2$ vectors in this style feature space as a new set of cluster centers $\left \{ \mathbf{c} _i \right \} _{i=1}^{K_2}$:
\vspace{-0.3em}
\begin{equation}
    \mathbf{c}_i\sim \mathit{U} \left ( \bm\mu _{ s}-\beta * \bm\sigma _{ s}, \bm\mu _{ s}+\beta * \bm\sigma _{ s} \right ), 
\end{equation}
where $\mathbf{c} _i\in \mathbb{R}^d  $ and $U$ represents uniform sampling.

\textbf{Cluster Sample Assignment.} After obtaining the cluster centers, we assign a fixed number of $Q$ samples that are closest to each cluster center to obtains all clusters. It is worth noting that during this process, a sample in $\mathcal{D}$ can be associated with multiple clusters, while some samples may not be assigned to any cluster at all. This differs from most traditional clustering methods, \textit{e.g.}, KMeans, where each sample can be assigned to a single cluster. Finally, we obtain a new dataset $\mathcal{D'}=\left \{\left \{ \left ( \boldsymbol{x}_{c}^i,\boldsymbol{y}_{c}^i ,c\right ) \right \} _{i=1}^Q\right \}_{c=1}^{K_2}$ for style prompt learning, where $c$ denotes pseudo style labels.

Finally, we combine the clusters obtained in Section 3.1 and those by our UPS. In this way, we obtain an MLLM training set $\mathcal{D''}=\mathcal{D}  \cup \mathcal{D'} $ with a total of ${K_1}+{K_2}$ style prompts for this combined training set.


\subsection{ReID Pre-training Dataset and Model}\label{sec:reid model}
As explained by Eq.\ref{Ins}, our HAM approach can be integrated with standard MLLMs to automatically generate desciptions of diverse styles. In this paper, we adopt the large-scale SYNTH-PEDES dataset \cite{zuo2023plip} as image source and create massive-scale pedestrian image-text pairs. We refer to our database as HAM-PEDES.

The same as \cite{tan2024harnessing, jiang2023cross}, we adopt the CLIP-ViT-B/16 \cite{radford2021learning} as the image encoder and a 12-layer transformer as the text encoder for the ReID model backbone. We feed the image and text into the image and text encoders, respectively, and obtain the holistic visual and text features. Following \cite{jiang2023cross,tan2024harnessing}, we adopt the Similarity Distribution Matching (SDM) loss to optimize the ReID model. The details of the SDM loss is provided in supplementary material.

%% file: sec/4_experiment.tex
\section{Experiments}
\label{sec:experiment}

\begin{table*}[tp]
\caption{Ablation study on the key components of our methods, \textit{i.e.}, HAM and UPS, in the direct transfer setting defined in \cite{tan2024harnessing}.}
\vspace{-1em}
\label{tab:ablation}

\centering
\resizebox{1.0\linewidth}{!}{
\begin{tabular}{l||cccccc||cccccc}
\hline
                                                     & \multicolumn{6}{c||}{LLaVA1.6-7B}                                                                                                                   & \multicolumn{6}{c}{Qwen-VL-Chat-7B}                                                                                                           \\ \cline{2-13} 
Method                                               & \multicolumn{2}{c|}{CUHK-PEDES}                 & \multicolumn{2}{c|}{ICFG-PEDES}                 & \multicolumn{2}{c||}{RSTPReid}                & \multicolumn{2}{c|}{CUHK-PEDES}                 & \multicolumn{2}{c|}{ICFG-PEDES}                 & \multicolumn{2}{c}{RSTPReid}                \\ \cline{2-13} 
                                                     & R1                   & \multicolumn{1}{c|}{mAP} & R1                   & \multicolumn{1}{c|}{mAP} & R1                   & mAP                   & R1                   & \multicolumn{1}{c|}{mAP} & R1                   & \multicolumn{1}{c|}{mAP} & R1                   & mAP                  \\ \hline \hline
\textbf{w/o fine-tuning}                             & \multicolumn{1}{l}{} & \multicolumn{1}{l|}{}    & \multicolumn{1}{l}{} & \multicolumn{1}{l|}{}    & \multicolumn{1}{l}{} & \multicolumn{1}{l||}{} & \multicolumn{1}{l}{} & \multicolumn{1}{l|}{}    & \multicolumn{1}{l}{} & \multicolumn{1}{l|}{}    & \multicolumn{1}{l}{} & \multicolumn{1}{l}{} \\
(1) static caption \cite{tan2024harnessing}                                     & 35.53                    & \multicolumn{1}{c|}{33.37}   & 16.90                    & \multicolumn{1}{c|}{9.11}   & 34.70                    & 27.33                     & 36.60                    & \multicolumn{1}{c|}{33.35}   & 21.38                    & \multicolumn{1}{c|}{9.59}   & 39.75                    & 27.35                    \\
(2) dynamic caption with 46 templates \cite{tan2024harnessing}                 & 41.50                    & \multicolumn{1}{c|}{38.38}   & 21.27                    & \multicolumn{1}{c|}{11.85}   & 38.65                    & 29.75                     & 39.73                    & \multicolumn{1}{c|}{36.09}   & 23.51                    & \multicolumn{1}{c|}{11.45}   & 39.95                    & 27.65                    \\
(3) dynamic caption with 6.8K templates          & 43.06                    & \multicolumn{1}{c|}{39.75}   & 21.97                    & \multicolumn{1}{c|}{12.02}   & 40.95                    & 31.46                     & 40.28                    & \multicolumn{1}{c|}{37.03}   & 24.01                    & \multicolumn{1}{c|}{12.22}   & 38.50                    & 28.54                    \\ \hline
\textbf{w/ fine-tuning}                              & \multicolumn{1}{l}{} & \multicolumn{1}{l|}{}    & \multicolumn{1}{l}{} & \multicolumn{1}{l|}{}    & \multicolumn{1}{l}{} & \multicolumn{1}{l||}{} & \multicolumn{1}{l}{} & \multicolumn{1}{l|}{}    & \multicolumn{1}{l}{} & \multicolumn{1}{l|}{}    & \multicolumn{1}{l}{} & \multicolumn{1}{l}{} \\
(4) Fine-tuning adaptor + 6.8K templates                                       & 47.06                    & \multicolumn{1}{c|}{45.07}   & 26.74                    & \multicolumn{1}{c|}{15.36}   & 36.60                    & 29.37                     & 42.99                    & \multicolumn{1}{c|}{39.98}   & 24.59                    & \multicolumn{1}{c|}{13.04}   & 38.55                    & 28.38                    \\
(5) HAM+random sample selection                & 52.87                    & \multicolumn{1}{c|}{47.24}   & 28.68                    & \multicolumn{1}{c|}{13.43}   & 43.35                    & 29.55                     & 43.60                    & \multicolumn{1}{c|}{39.45}   & 21.87                    & \multicolumn{1}{c|}{10.17}   & 42.15                    & 29.24                    \\
(6) HAM+KMeans \cite{macqueen1967some} ($K_1$=1000)                         & 55.28                    & \multicolumn{1}{c|}{49.99}   & 31.08                    & \multicolumn{1}{c|}{16.39}   & 45.65                    & 33.71                     & 52.19                    & \multicolumn{1}{c|}{46.59}   & 28.26                    & \multicolumn{1}{c|}{13.66}   & 44.65                    & 32.35                    \\
(7) HAM+DBSCAN \cite{ester1996density}                         & 48.02                    & \multicolumn{1}{c|}{44.25}   & 25.75                    & \multicolumn{1}{c|}{13.98}   & 41.95                    & 31.12                     & 46.93                    & \multicolumn{1}{c|}{42.08}   & 24.92                    & \multicolumn{1}{c|}{11.97}   & 41.40                    & 29.29                    \\
(8) HAM+UPS ($K_2$=1000)                             & 59.05                    & \multicolumn{1}{c|}{52.92}   & 35.06                    & \multicolumn{1}{c|}{18.34}   & 50.90                    & 36.94                     & 54.53                    & \multicolumn{1}{c|}{48.34}   & 30.57                    & \multicolumn{1}{c|}{14.80}   & 45.40                    & 32.24                    \\
(9) HAM+DBSCAN+UPS ($K_2$=1000)                             & 57.71                    & \multicolumn{1}{c|}{51.46}   & 34.97                    & \multicolumn{1}{c|}{18.38}   & 48.10                    & 35.61                     & 53.56                    & \multicolumn{1}{c|}{47.95}   & 29.57                    & \multicolumn{1}{c|}{14.93}   & 44.84                    & 32.20                    \\
(10) HAM+KMeans ($K_1$=2000)                        & 56.02                    & \multicolumn{1}{c|}{50.61}   & 31.09                    & \multicolumn{1}{c|}{16.24}   & 45.45                    & 33.93                     & 51.20                    & \multicolumn{1}{c|}{45.57}   & 28.34                    & \multicolumn{1}{c|}{12.88}   & 45.40                    & 31.75                    \\
(11) HAM+UPS ($K_2$=2000)                          & 59.34                    & \multicolumn{1}{c|}{52.30}   & 35.40                    & \multicolumn{1}{c|}{18.61}   & 49.20                    & 36.37                     & 54.49                    & \multicolumn{1}{c|}{48.28}   & 30.70                    & \multicolumn{1}{c|}{14.58}   & 45.40                    & 31.34                    \\
(12) HAM+KMeans+UPS ($K_1$=2000,$K_2$=2000)                          & 59.92                    & \multicolumn{1}{c|}{53.62}   & 36.09                    & \multicolumn{1}{c|}{18.86}   & 50.75                    & 36.89                     & 54.72                    & \multicolumn{1}{c|}{48.61}   & 30.14                    & \multicolumn{1}{c|}{14.78}   & 44.30                    & 31.83                    \\
(13) HAM+KMeans+UPS ($K_1$=1000,$K_2$=1000) & \textbf{60.60}                    & \multicolumn{1}{c|}{\textbf{54.32}}   & \textbf{36.87}                    & \multicolumn{1}{c|}{\textbf{19.33}}   & \textbf{51.75}                    & \textbf{38.02}                     & \textbf{55.34}                    & \multicolumn{1}{c|}{\textbf{48.91}}   & \textbf{31.89}                    & \multicolumn{1}{c|}{\textbf{15.45}}   & \textbf{46.05}                    & \textbf{32.71}                    \\ \hline
\end{tabular}}
\vspace{-1em}
\end{table*}

\subsection{Datasets and Metrics}
\textbf{CUHK-PEDES}
\cite{li2017person} is the first dataset for the text-to-image ReID task. It contains a total of 40,206 images from 13,003 identities. Each image was annotated by two different annotators. We follow its official data spliting protocol, where the training set comprises 34,054 images from 11,003 identities, the validation set contains 3,078 images from 1,000 identities, and the testing set contains 3,074 images from 1,000 identities, respectively.

\textbf{ICFG-PEDES} \cite{ding2021semantically} consists of 54,552 images from 4,102 identities. Each image owns one caption. Its training set comprises 34,674 image-text pairs from 3,102 identities, while testing set includes the remaining 19,848 image-text pairs from 1,000 identities.

\textbf{RSTPReid} \cite{zhu2021dssl} contains 20,505 images from 4,104 identities across 15 different cameras. Each identity has 5 images from different cameras, and each image is paired with 2 captions. According to the official data spliting protocol, the training set includes image-text pairs from 3,701 identities, while both the validation and testing sets contain 200 identities, respectively.

\textbf{SYNTH-PEDES} \cite{zuo2023plip} is another large-scale dataset with automatically generated textual annotations. It contains 4,791,711 pedestrian images from 312,321 identities. These images are collected from the LUPerson-NL \cite{gong2017look} and LPW \cite{song2018region} datasets. Each image has an average of 2.53 captions generated by an image captioner \cite{zuo2023plip}. Our HAM-PEDES database is based on the 1.0 million pedestrian images randomly selected from SYNTH-PEDES.

\textbf{Evaluate Metrics.} Following existing works \cite{jiang2023cross,shao2023unified}, we adopt the Rank-\textit{k} (\textit{k}=1,5,10) and mean Average Precision (mAP) as the evaluation metrics for above benchmarks. Higher Rank-$k$ and mAP indicate better performance.

\subsection{Implementation Details}
To demonstrate the wide suitability of our HAM approach, we conduct ablation studies on two representative MLLMs, \textit{i.e.}, LLaVA1.6-7B \cite{liu2024llava} and Qwen-VL-Chat-7B \cite{bai2023qwen}, respectively. The latter was also adopted in a recent work for MLLM-based text-to-image ReID \cite{tan2024harnessing}. In HAM, each cluster in UPS contains $Q$ = 200 samples. The length of the learnable prompt is $M$ = 10. The hyper-parameter $\beta$ is set to 7. More details about hyper-parameter tuning are provided in the supplementary material. We train the style prompts and fine-tune the adapter of the Qwen model with the AdamW optimizer \cite{loshchilov2017decoupled} for 2 epochs and a learning rate of 2e-4. For the LLaVA1.6-7B model, we adopt a larger learning rate of 4e-4. We conduct experiments on 8 GeForce RTX 3090 GPUs.

For ReID model training, we resize image to 384$\times $128 pixels and apply random horizontal flipping, random cropping, and random erasing as image data augmentation. The maximum length of text tokens is set to 77. The ReID model is trained using the Adam optimizer with a learning rate of 1e-5 and a cosine learning rate decay strategy. We train each model on 8 GPUs, with each one processing 128 images. The training process lasts for 30 epochs.

\subsection{Ablation Study}
In this subsection, we employ the training set of CUHK-PEDES \cite{li2017person} to train the style prompts and fine-tune the adaptor in MLLM. Then, we adopt similar evaluation settings as \cite{tan2024harnessing} to facilitate fair comparison with recent works. Specifically, we adopt two MLLMs equipped with HAM to annotate 0.1 million images with one caption, respectively. These images are randomly sampled from the SYNTH-PEDES database. The number of clusters in HAM is set to $K_1$ = 1,000 and $K_2$ = 1,000. We adopt the obtained image-text pairs to train the ReID model described in Section \ref{sec:reid model} and directly test the its performance without extra fine-tuning on three popular databases (\textit{i.e.}, direct transfer setting \cite{tan2024harnessing}). Experimental results are shown in Table \ref{tab:ablation}.

\textbf{Effectiveness of HAM.} First, we conduct three experiments in Table \ref{tab:ablation} to show the performance of traditional methods that rely on description templates to enhance the output diversity of MLLMs. In (1), we do not adopt any template and employ fixed instruction to prompt MLLM to produce a textual description for each image. The obtained caption is called ``static caption" in \cite{tan2024harnessing} and our Table \ref{tab:ablation}. In (2), we randomly pick one of 46 templates from \cite{tan2024harnessing} to enhance the output diversity of MLLMs. The obtained description is called ``dynamic caption" \cite{tan2024harnessing}. In (3), we try to explore the extreme performance of the template-based methods. We extract 68,126 templates according to all human annotations in the training set of CUHK-PEDES and adopt the same way as (2) to obtain ``dynamic caption". The details to obtain these templates are provided in supplementary material. Results in Table \ref{tab:ablation} show that the increase in the number of templates yields only marginal ReID performance improvements. This maybe due to two reasons: there is rich redundancy in the obtained templates; a portion of style information is lost in the manually designed process to generate templates.

To demonstrate the effectiveness of HAM, we further conduct three experiments from (4) to (6) in Table \ref{tab:ablation}. In (4), we fine-tune parameters of the vision-language adaptor based on (3) and we do not introduce the learnable style prompts. In (5), we fine-tune both 1,000 style prompts and the vision-language adaptor. We randomly sample 200 image-text pairs from the training set of CUHK-PEDES for each style prompt. In (6), we obtain the training data for each style prompt according to the method described in Section \ref{sec:HAM} and we adopt KMeans clustering, resulting in 1,000 style prompts. We then randomly pick one style prompt to generate a caption for each image. As shown in Table 1, (6) significantly outperforms both (4) and (5). For example, the Rank-1 accuracy of CUHK-PEDES surpass (4) and (5) by 9.2\% and 8.59\% in Qwen, respectively. These experimental results indicate that HAM effectively models the styles of different human annotators; therefore, it significantly enhances the diversity of MLLM-generated captions.

\textbf{Effectiveness of UPS.} We compare the performance of three clustering approaches, \textit{i.e.}, KMeans \cite{macqueen1967some}, DBSCAN \cite{ester1996density}, and our UPS, in the framework of HAM. These experiments are denoted as (6) to (8) in Table 1. Since DBSCAN is sensitive to density of sample distribution, it tends to classify most sparsely distributed style features as noise samples, making it unsuitable for our HAM. Details for DBSCAN clustering are provided in the supplementary material. Compared with (6) and (7), our UPS method outperforms KMeans and DBSCAN by 3.77\% and 11.03\% in Rank-1 accuracy of CUHK-PEDES with LLaVA1.6, respectively. This indicates that UPS captures a broader range of human annotator styles within the style feature space.

\textbf{Combination of UPS and KMeans.} The KMeans and UPS complement with each other. KMeans focuses on areas with dense samples while UPS is irrelevant to sample density. Therefore, we combine the obtained clusters in both KMeans and UPS in (13), resulting in a total of 2,000 clusters for style modeling. Compare with (10) and (11), where KMeans and UPS produce 2,000 clusters, respectively, (13) achieves the best results.  These results justify our conjecture that UPS and KMeans are complementary for our HAM framework. Finally, with the LLaVA1.6 model, our complete method (13) achieves a significant improvement in the Rank-1 accuracy by 10.13\% and 15.15\% on ICFG-PEDES and RSTPReid compared to (4), even though it only learns the styles of human annotations from the CUHK-PEDES. These results also demonstrate that our method produces diverse captions that generalize to both ICFG-PEDES and RSTPReid databases.




\begin{table}[tp]
	\caption{Comparisons with existing pre-training datasets under the direct transfer setting defined in \cite{tan2024harnessing}. There are 2.53, 4.0, and 2.0 captions per image on average for \cite{zuo2023plip}, \cite{tan2024harnessing}, and ours, respectively.}
	\vspace{-1em}
	\label{tab:zero-shot}
	\centering
	\resizebox{0.45\textwidth}{!}{%
		\begin{tabular}{c|cc|cc|cc}
			\toprule[1pt]
			\multirow{2}{*}{Pre-training Dataset} & \multicolumn{2}{c|}{CUHK-PEDES} & \multicolumn{2}{c|}{ICFG-PEDES} & \multicolumn{2}{c}{RSTPReid} \\ \cline{2-7} 
			& R1 & mAP & R1 & mAP & R1 & mAP \\ \hline \hline
			None & 12.65 & 11.15 & 6.67 & 2.51 & 13.45 & 10.31 \\ \hline
			MALS(1.5M) \cite{yang2023towards} & 19.36 & 18.62 & 7.93 & 3.52 & 22.85 & 17.11 \\ \hline
			LuPerson-T(0.95M) \cite{shao2023unified} & 21.88 & 19.96 & 11.46 & 4.56 & 22.40 & 17.08 \\ 
            \hline
            SYNTH-PEDES(1.0M) \cite{zuo2023plip} & 57.58  & 52.45 &57.08  &32.06  & 42.69 & 31.18 \\ 
			\hline
			LuPerson-MLLM(1.0M) \cite{tan2024harnessing} & 57.61  & 51.44 & 38.36   & 20.43  & 51.50  &  37.34  \\
   \hline
			Ours (0.1 M) & 60.74  & 54.57  & 50.96  &28.00  &49.80   &36.97  \\
			Ours (1.0 M) & \textbf{70.15}  & \textbf{62.77}  & \textbf{59.63}  &\textbf{34.92}  &\textbf{58.85}   &\textbf{44.11}   \\ \hline
	\end{tabular}}
	\vspace{0em}
\end{table}

\begin{table}[tp]
\caption{Comparisons with existing pre-training datasets under the traditional pre-training and fine-tuning setting.}
	\vspace{-1em}
	\label{tab:finetune}
	\centering
	\resizebox{0.45\textwidth}{!}{%
\begin{tabular}{c|cl|cl|cl}
\hline
\toprule[1pt]\multirow{2}{*}{Init Parameters}  & \multicolumn{2}{c|}{CUHK-PEDES}  & \multicolumn{2}{c|}{ICFG-PEDES}  & \multicolumn{2}{c}{RSTPReid}     \\ \cline{2-7} 
                                  & R1                   & mAP       & R1                   & mAP       & R1                   & mAP       \\ \hline \hline
CLIP                              & 73.48                & 66.21     & 63.83                & 38.37     & 60.40                & 47.70     \\ \hline
MALS(1.5M) \cite{yang2023towards}       & 74.05                & 66.57     & 64.37                & 38.85     & 61.90                & 48.08     \\ \hline
LuPerson-T(0.95M) \cite{shao2023unified} & 74.37                & 66.60     & 64.50                & 38.22     & 62.20                & 48.33     \\ \hline
SYNTH-PEDES(1.0M)                & 74.25                & 66.52     & 65.79                & 39.43     & 64.35                & 50.93     \\ \hline
LuPerson-MLLM(1.0M)                & 76.82                & 69.55     & 67.05                & 41.51     & 68.50                & 53.20     \\ \hline
Ours(1.0M)                        & \textbf{77.71}            & \textbf{69.68} & \textbf{68.25}                     & \textbf{42.30}          & \textbf{71.69}            & \textbf{55.19} \\ \hline
\end{tabular}}
\vspace{-1.5em}
\end{table}


\begin{table*}[tp]
\caption{Comparisons with state-of-the-art ReID methods under the traditional evaluation setting.}
\vspace{-0.8em}
\label{tab:sota-traditional}

\centering
\resizebox{0.975\linewidth}{!}{
\begin{tabular}{lcccccccccccccc}
\hline
\multicolumn{1}{l|}{\multirow{2}{*}{Method}}                  & \multirow{2}{*}{Image Enc.} & \multicolumn{1}{c|}{\multirow{2}{*}{Text Enc.}} & \multicolumn{4}{c|}{CUHK-PEDES}                                                                                & \multicolumn{4}{c|}{ICFG-PEDES}                                                                                & \multicolumn{4}{c}{RSTPReid}                                                                                  \\ \cline{4-15} 
\multicolumn{1}{l|}{}                                         &                             & \multicolumn{1}{c|}{}                           & R1                        & R5                        & R10                       & \multicolumn{1}{c|}{mAP}   & R1                        & R5                        & R10                       & \multicolumn{1}{c|}{mAP}   & R1                        & R5                        & R10                       & mAP                       \\ \hline
\multicolumn{6}{l}{\textit{with CLIP \cite{radford2021learning} backbone:}}                                                                                                                                                       &                            &                           &                           &                           &                            &                           &                           &                           &                           \\ \hline
\multicolumn{1}{l|}{Han et al. \cite{han2021text}}            & CLIP-RN101                  & \multicolumn{1}{c|}{CLIP-Xformer}               & 64.08                     & 81.73                     & 88.19                     & \multicolumn{1}{c|}{60.08} & -                         & -                         & -                         & \multicolumn{1}{c|}{-}     & -                         & -                         & -                         & -                         \\
\multicolumn{1}{l|}{IRRA \cite{jiang2023cross}}               & CLIP-ViT                    & \multicolumn{1}{c|}{CLIP-Xformer}               & 73.38                     & 89.93                     & 93.71                     & \multicolumn{1}{c|}{66.10} & 63.46                     & 80.25                     & 85.82                     & \multicolumn{1}{c|}{38.06} & 60.20                     & 81.30                     & 88.20                     & 47.17                     \\
\multicolumn{1}{l|}{TBPS-CLIP \cite{cao2024empirical}}                                & CLIP-ViT                    & \multicolumn{1}{c|}{CLIP-Xformer}               & \multicolumn{1}{l}{73.54} & \multicolumn{1}{l}{88.19} & \multicolumn{1}{l}{92.35} & \multicolumn{1}{l|}{65.38} & \multicolumn{1}{l}{65.05} & \multicolumn{1}{l}{80.34} & \multicolumn{1}{l}{85.47} & \multicolumn{1}{l|}{39.83} & \multicolumn{1}{l}{61.95} & \multicolumn{1}{l}{83.55} & \multicolumn{1}{l}{88.75} & \multicolumn{1}{l}{48.26} \\
\multicolumn{1}{l|}{CFAM \cite{zuo2024ufinebench}}                                     & CLIP-ViT                    & \multicolumn{1}{c|}{CLIP-Xformer}               & \multicolumn{1}{l}{73.67} & \multicolumn{1}{l}{89.71} & \multicolumn{1}{l}{93.57} & \multicolumn{1}{l|}{65.94} & \multicolumn{1}{l}{63.57} & \multicolumn{1}{l}{80.57} & \multicolumn{1}{l}{86.32} & \multicolumn{1}{l|}{38.34} & \multicolumn{1}{l}{60.51} & \multicolumn{1}{l}{82.85} & \multicolumn{1}{l}{89.71} & \multicolumn{1}{l}{47.64} \\
\multicolumn{1}{l|}{UMSA \cite{zhao2024unifying}}                                     & CLIP-ViT                    & \multicolumn{1}{c|}{CLIP-Xformer}               & \multicolumn{1}{l}{74.25} & \multicolumn{1}{l}{89.83} & \multicolumn{1}{l}{93.58} & \multicolumn{1}{c|}{66.15}     & 65.62                         & 80.54                         & 85.83                         & \multicolumn{1}{c|}{38.78}     & 63.40                         & 83.30                         & 90.30                         & 49.28                         \\
\multicolumn{1}{l|}{LSPM \cite{li2024learning}}                                     & CLIP-ViT                    & \multicolumn{1}{c|}{CLIP-Xformer}               & \multicolumn{1}{l}{74.38} & \multicolumn{1}{l}{89.51} & \multicolumn{1}{l}{93.42} & \multicolumn{1}{c|}{67.74} & 64.40                     & 79.96                     & 85.41                     & \multicolumn{1}{c|}{42.60} & -                         & -                         & -                         & -                         \\
\multicolumn{1}{l|}{IRLT \cite{liu2024causality}}                                     & CLIP-ViT                    & \multicolumn{1}{c|}{CLIP-Xformer}               & \multicolumn{1}{l}{74.46} & \multicolumn{1}{l}{90.19} & \multicolumn{1}{l}{94.01} & \multicolumn{1}{c|}{-}     & 64.72                     & 81.35                     & 86.31                     & \multicolumn{1}{c|}{-}     & 61.49                     & 82.26                     & 89.23                     & -                         \\
\multicolumn{1}{l|}{MDRL \cite{yang2024multi}}                                     & CLIP-ViT                    & \multicolumn{1}{c|}{CLIP-Xformer}               & \multicolumn{1}{l}{74.56} & \multicolumn{1}{l}{92.56} & \multicolumn{1}{l}{96.30} & \multicolumn{1}{c|}{-}     & 65.88                          &85.25                           &90.38                           & \multicolumn{1}{c|}{-}      &  -                         &  -                         &    -                       & -                          \\
\multicolumn{1}{l|}{FSRL \cite{wang2024fine}}                                     & CLIP-ViT                    & \multicolumn{1}{c|}{CLIP-Xformer}               & \multicolumn{1}{l}{74.65} & \multicolumn{1}{l}{89.77} & \multicolumn{1}{l}{94.03} & \multicolumn{1}{c|}{67.49} & 64.01                     & 80.42                     & 85.56                     & \multicolumn{1}{c|}{39.64} & 60.20                     & 81.40                     & 88.60                     & 47.38                     \\
\multicolumn{1}{l|}{Propot \cite{yan2024prototypical}}                                   & CLIP-ViT                    & \multicolumn{1}{c|}{CLIP-Xformer}               & \multicolumn{1}{l}{74.89} & \multicolumn{1}{l}{89.90} & \multicolumn{1}{l}{94.17} & \multicolumn{1}{c|}{67.12} & 65.12                     & 81.57                     & 86.97                     & \multicolumn{1}{c|}{42.93} & 61.87                     & 83.63                     & 89.70                     & 47.82                     \\
\multicolumn{1}{l|}{SAP-SAM \cite{wang2024fine}}                                  & CLIP-ViT                    & \multicolumn{1}{c|}{CLIP-Xformer}               & \multicolumn{1}{l}{75.05} & \multicolumn{1}{l}{89.93} & \multicolumn{1}{l}{93.73} & \multicolumn{1}{c|}{-}     & 63.97                     & 80.84                     & 86.17                     & \multicolumn{1}{c|}{-}     & 62.85                     & 82.65                     & 89.85                     & -                         \\
\multicolumn{1}{l|}{PLOT \cite{park2025plot}}                                     & CLIP-ViT                    & \multicolumn{1}{c|}{CLIP-Xformer}               & \multicolumn{1}{l}{75.28} & \multicolumn{1}{l}{90.42} & \multicolumn{1}{l}{94.12} & \multicolumn{1}{c|}{-}     & 65.76                     & 81.39                     & 86.73                     & \multicolumn{1}{c|}{-}     & 61.80                     & 82.85                     & 89.45                     & -                         \\
\multicolumn{1}{l|}{RDE \cite{qin2024noisy}}                                      & CLIP-ViT                    & \multicolumn{1}{c|}{CLIP-Xformer}               & \multicolumn{1}{l}{75.94} & \multicolumn{1}{l}{90.14} & \multicolumn{1}{l}{94.12} & \multicolumn{1}{l|}{67.56} & 67.68                     & 82.47                     & 87.36                     & \multicolumn{1}{c|}{40.06}     & 65.35                     & 83.95                     & 89.90                     & 50.88                         \\ \hline
\textit{with ALBEF \cite{li2021align} backbone:}              &                             &                                                 &                           &                           &                           &                            &                           &                           &                           &                            &                           &                           &                           &                           \\ \hline
\multicolumn{1}{l|}{APTM \cite{yang2023towards}}              & Swin-B                      & \multicolumn{1}{c|}{BERT-base}                  & 76.53                     & 90.04                     & 94.15                     & \multicolumn{1}{c|}{66.91} & 68.51                     & 82.99                     & 87.56                     & \multicolumn{1}{c|}{41.22} & 67.50                     & 85.70                     & 91.45                     & 52.56                     \\
\multicolumn{1}{l|}{RaSa \cite{bai2023rasa}}                  & CLIP-ViT                    & \multicolumn{1}{c|}{BERT-base}                  & 76.51                     & 90.29                     & 94.25                     & \multicolumn{1}{c|}{69.38} & 65.28                     & 80.40                     & 85.12                     & \multicolumn{1}{c|}{41.29} & 66.90                     & 86.50                     & 91.35                     & 52.31                     \\
\multicolumn{1}{l|}{AUL \cite{li2024adaptive}}                  & Swin-B                    & \multicolumn{1}{c|}{BERT-base}                  & 77.23                     & 90.43                     & 94.25                     & \multicolumn{1}{c|}{-} & 69.16                     & 83.32                    & 88.37                     & \multicolumn{1}{c|}{-} & 71.65                     & 87.55                     & 92.05                     & -                     \\ \hline
\multicolumn{6}{l}{\textit{with CLIP \cite{radford2021learning} backbone:}}                                                                                                                                                       &                            &                           &                           &                           &                            &                           &                           &                           &                           \\ \hline
\multicolumn{1}{l|}{MALS \cite{yang2023towards} + IRRA \cite{jiang2023cross}}       & CLIP-ViT                    & \multicolumn{1}{c|}{CLIP-Xformer}               & 74.05                     & 89.48                     & 93.64                     & \multicolumn{1}{c|}{66.57} & 64.37                     & 80.75                     & 86.12                     & \multicolumn{1}{c|}{38.85} & 61.90                     & 80.60                     & 89.30                     & 48.08                     \\
\multicolumn{1}{l|}{LuPerson-T \cite{shao2023unified} + IRRA \cite{jiang2023cross}} & CLIP-ViT                    & \multicolumn{1}{c|}{CLIP-Xformer}               & 74.37                     & 89.51                     & 93.97                     & \multicolumn{1}{c|}{66.60} & 64.50                     & 80.24                     & 85.74                     & \multicolumn{1}{c|}{38.22} & 62.20                     & 83.30                     & 89.75                     & 48.33                     \\
\multicolumn{1}{l|}{SYNTH-PEDES(1.0 M) \cite{zuo2023plip}+ IRRA \cite{jiang2023cross}}                 & CLIP-ViT                    & \multicolumn{1}{c|}{CLIP-Xformer}               & \multicolumn{1}{l}{74.25}      & \multicolumn{1}{l}{89.49}      & \multicolumn{1}{l}{93.68}      & \multicolumn{1}{l|}{66.52}      & \multicolumn{1}{l}{65.79}      & \multicolumn{1}{l}{81.94}      & \multicolumn{1}{l}{87.32}      & \multicolumn{1}{l|}{39.43}      & \multicolumn{1}{l}{64.35}      & \multicolumn{1}{l}{83.75}      & \multicolumn{1}{l}{91.00}      & \multicolumn{1}{l}{50.93}      \\
\multicolumn{1}{l|}{LUPeron-MLLM(1.0 M) \cite{tan2024harnessing} + IRRA \cite{jiang2023cross}}               & CLIP-ViT                    & \multicolumn{1}{c|}{CLIP-Xformer}               & 76.82                     & 91.16                     & 94.46                     & \multicolumn{1}{c|}{69.55} & 67.05                     & 82.16                     & 87.33                     & \multicolumn{1}{c|}{41.51} & 68.50                     & 87.15                     & 92.10                     & 53.02                     \\
\multicolumn{1}{l|}{HAM-PEDES(1.0 M) + IRRA \cite{jiang2023cross}}                  & CLIP-ViT                    & \multicolumn{1}{c|}{CLIP-Xformer}               & \multicolumn{1}{l}{77.71}      & \multicolumn{1}{l}{\textbf{91.42}}      & \multicolumn{1}{l}{94.57}      & \multicolumn{1}{l|}{69.68}      & \multicolumn{1}{l}{68.25}      & \multicolumn{1}{l}{83.30}      & \multicolumn{1}{l}{88.15}      & \multicolumn{1}{l|}{42.30}      & \multicolumn{1}{l}{71.69}      & \multicolumn{1}{l}{\textbf{87.85}}      & \multicolumn{1}{l}{\textbf{93.30}}      & \multicolumn{1}{l}{55.19}      \\
\multicolumn{1}{l|}{HAM-PEDES(1.0 M) + RDE \cite{qin2024noisy}}                   & CLIP-ViT                    & \multicolumn{1}{c|}{CLIP-Xformer}               & \multicolumn{1}{l}{\textbf{77.99}}      & \multicolumn{1}{l}{91.34}      & \multicolumn{1}{l}{\textbf{95.03}}      & \multicolumn{1}{l|}{\textbf{69.72}}      & \multicolumn{1}{l}{\textbf{69.95}}      & \multicolumn{1}{l}{\textbf{83.88}}      & \multicolumn{1}{l}{\textbf{88.39}}      & \multicolumn{1}{l|}{\textbf{42.72}}      & \multicolumn{1}{l}{\textbf{72.50}}      & \multicolumn{1}{l}{87.70}      & \multicolumn{1}{l}{91.95}      & \multicolumn{1}{l}{\textbf{55.47}}      \\ \hline
\vspace{-3em}
\end{tabular}}
\end{table*}

\subsection{Comparisons with State-of-the-Art Methods} 
In this section, we employ only LLaVA1.6 for dataset annotation as it demonstrates better performance. To facilitate fair comparison with a very recent work \cite{zuo2023plip}, we combine the training sets of CUHK-PEDES and ICFG-PEDES to train the style prompts and fine-tune the language-vision adaptor of the LLaVA. It is worth noting we have removed the training data from ICFG-PEDES for the identities that also appear in the testing set of RSTPReid. Since the training set for HAM becomes larger, we empirically increase the number of clusters to 1,500 for both KMeans and UPS. Therefore, we train a total of 3,000 style prompts. Finally, we annotate each image with two captions by our HAM method and name the obtained database with 1.0M images as HAM-PEDES.

\begin{figure}[t]
  \centerline{\includegraphics[width=1.0\linewidth]{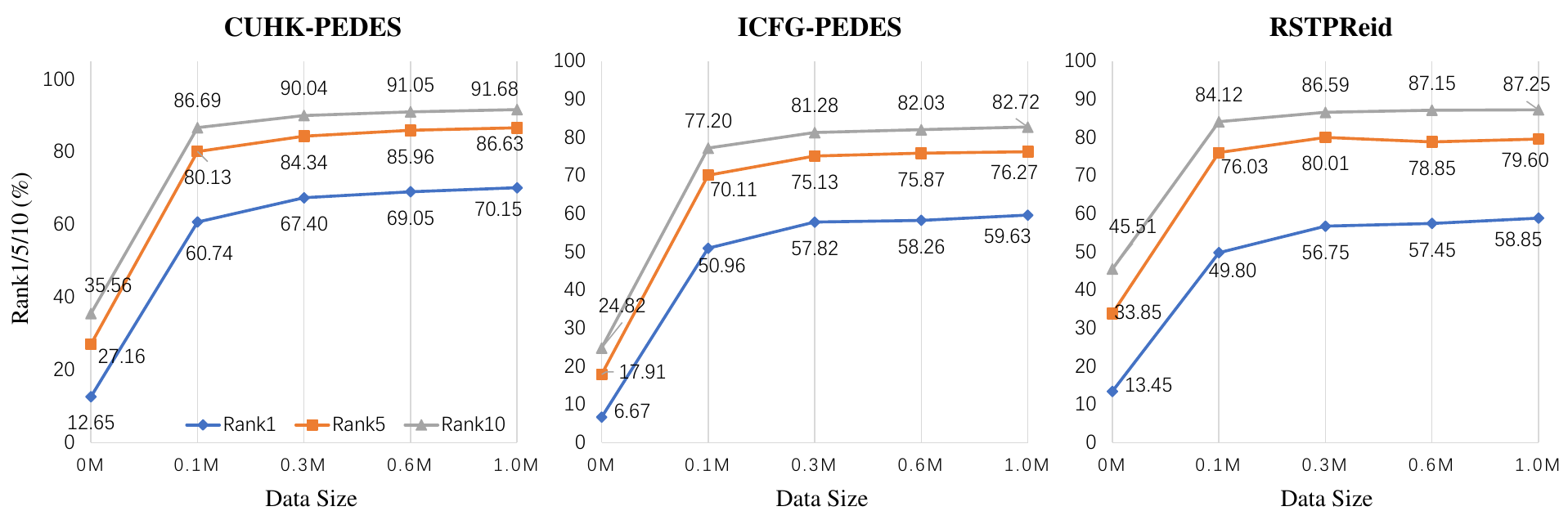}}
  \vspace{-1em}
  \caption{Pre-training data size’s impact on direct transfer ReID performance. Best viewed with zoom-in.}
  \label{fig:datasize}
  \vspace{-1.5em}
\end{figure}

\textbf{Impact of Pre-training Data Size.} The size of pre-training data affects the model's generalizability. We gradually increase the data size from 0.1M, 0.3M, 0.6M, to 1.0M images for model pre-training and evaluate the direct transfer ReID performance across three benchmarks. As shown in Figure \ref{fig:datasize}, the ReID performance steadily improves with the increase in data volume. Finally, when the data size increases to 1.0M, the Rank-1 accuracy of the RSTPReid improves by 9.05\% over that of 0.1M, demonstrating that our approach’s effectiveness in scaling to large dataset.

\textbf{Comparisons with Other Pre-training Datasets.} In this experiment, we compare the performance of ReID models with five large-scale pre-training datasets: MALS \cite{yang2023towards}, LUPerson-T \cite{shao2023unified}, LUPerson-MLLM \cite{tan2024harnessing}, SYNTH-PEDES \cite{zuo2023plip}, and our HAM-PEDES. MALS includes 1.5M images and their captions are produced by BLIP2. LUPerson-T consists of 0.95M images and it utilizes 456 templates to enhance the diversity of the captions. Similarly, LUPerson-MLLM employs 46 templates to enhance captioning diversity of two MLLMs. It comprises 1.0M images, with each image annotated with 4.0 captions by two MLLMs. SYNTH-PEDES contains 4.7M images, each annotated with an average of 2.53 captions generated by a captioner trained on the CUHK-PEDES and ICFG-PEDES datasets. To facilitate fair comparison, we adopt the same 1.0M images as HAM-PEDES for the experiments on SYNTH-PEDES. Compared with \cite{tan2024harnessing,zuo2023plip}, we annotate the least number of captions per image.

We utilize each of the aforementioned datasets to train the ReID model according to the description in Section \ref{sec:reid model}. Then, we evaluate the model’s ReID performance under the direct transfer and fine-tuning settings. The experimental results are summarized in Tables \ref{tab:zero-shot} and \ref{tab:finetune}, respectively. Table \ref{tab:zero-shot} shows that the model pre-trained on our HAM-PEDES dataset achieves superior direct transfer performance across three benchmarks. Specifically, the Rank-1 accuracies on the CUHK-PEDES, ICFG-PEDES and RSTPReid attain remarkable 70.15\%, 59.63\% and 58.85\%, respectively and it surpasses all of the other pre-training datasets.

Table \ref{tab:finetune} provides the comparisons of the fine-tuning performance. Following \cite{tan2024harnessing}, we adopt the IRRA method \cite{jiang2023cross} during the fine-tuning stage and initialize its parameters using the model pretrained on different pre-training datasets. The results show that the model pretrained on HAM-PEDES achieves optimal performance after fine-tuning on downstream datasets. Specifically, the Rank-1 accuracy and mAP on RSTPReid outperform LuPerson-MLLM by 3.19\% and 1.99\%, respectively, even though the HAM-PEDES has only half the number of captions compared to LUPerson-MLLM and is only derived from a single MLLM.


\textbf{Comparisons with State-of-the-Art ReID Methods.} The comparisons are shown in Table \ref{tab:sota-traditional}. We observe that after fine-tuning with the IRRA \cite{jiang2023cross} method, our pre-trained model achieves superior Rank-1 performance compared to all previous methods, even outperforms those that adpot more powerful but complicated backbones (\textit{i.e.}, ALBEF). It is worth noting the average retrieval speeds for AUL \cite{li2024adaptive} and ours are 123ms and 1.5ms on a 3090 GPU for each textual description on the CUHK-PEDES, respectively. Moreover, we also adopt the RDE \cite{qin2024noisy} model that is also based on the CLIP backbone for fine-tuning. The results show that the Rank-1 accuracy of RDE is improved by 2.05\%, 2.27\% and 7.20\% on CUHK-PEDES, ICFG-PEDES, and RSTPReid, respectively, which achieves the new SOTA performance. These experimental results further justify the effectiveness of our HAM-PEDES pre-training dataset.

%% file: sec/5_conclusion.tex
\section{Conclusion and Limitations}

This paper explores the challenge of limited diversity in description styles for MLLM-based annotations in text-to-image ReID. To address this, we propose a Human Annotator Modeling (HAM) approach that enables MLLMs to mimic the description styles of thousands of human annotators. Moreover, we introduce the Uniform Prototype Sampling (UPS) strategy to fully explore the human annotation styles and further enhance the diversity of MLLM-generated annotations. Finally,  we create a large-scale database for text-to-image ReID named HAM-PEDES. Extensive experiments demonstrate that our methods significantly enhance the annotation diversity in HAM-PEDES, and further enables us to achieve state-of-the-art ReID performance on popular benchmarks. Our work also has limitations: it does not account for the impact of noise in the obtained annotations. In the future, we will develop methods that relieve the hallucination of MLLMs. \\
\textbf{Broader Impacts.} Our approach significantly enhances the style diversity in MLLM-annotated descriptions. It has broad application potentials to reduce human labeling efforts in database construction. To the best of our knowledge, our work does not have obvious negative social impacts. \\
\textbf{Acknowledgement.} This work was supported by the National Natural Science Foundation of China (Grant 62476099, 62076101), Guangdong Basic and Applied Basic Research Foundation (Grant 2024B1515020082, 2023A1515010007), the Guangdong Provincial Key Laboratory of Human Digital Twin (Grant 2022B1212010004), the TCL Young Scholars Program, and the 2024 Tencent AI Lab Rhino-Bird Focused Research Program.
